\pdfoutput=1

\documentclass[11pt]{article}

\usepackage{acl}

\usepackage{times}
\usepackage{latexsym}
\usepackage{graphicx}
\usepackage{booktabs} 
\usepackage{tcolorbox}
\usepackage{hyperref}

\usepackage[T1]{fontenc}

\usepackage[utf8]{inputenc}

\usepackage{microtype}

\title{Low-code LLM: Graphical User Interface over Large Language Models}

\author{Yuzhe Cai\thanks{\ \ The first two authors contributed equally. This work was performed during the first author’s internship at Microsoft Research Asia} , Shaoguang Mao$^*$, Wenshan Wu, Zehua Wang, Yaobo Liang, Tao Ge\\
{\bf Chenfei Wu, Wang You, Ting Song, Yan Xia, Nan Duan \and Furu Wei}\\
Microsoft Research Asia\\
\texttt{v-yuzhecai, shaoguang.mao, wenswu, zehwang, yalia, tage,} \\
\texttt{chewu, v-wangyou, tsong, yanxia, nanduan, fuwei@microsoft.com}}

\begin{document}
\maketitle
\begin{abstract}
Utilizing Large Language Models (LLMs) for complex tasks is challenging, often involving a time-consuming and uncontrollable prompt engineering process. This paper introduces a novel human-LLM interaction framework, \textbf{Low-code LLM}. It incorporates six types of simple low-code visual programming interactions to achieve more controllable and stable responses. Through visual interaction with a graphical user interface, users can incorporate their ideas into the process without writing trivial prompts. 
The proposed Low-code LLM framework consists of a Planning LLM that designs a structured planning workflow for complex tasks, which can be correspondingly edited and confirmed by users through low-code visual programming operations, and an Executing LLM that generates responses following the user-confirmed workflow. 
We highlight three advantages of the low-code LLM: user-friendly interaction, controllable generation, and wide applicability. We demonstrate its benefits using four typical applications. 
By introducing this framework, we aim to bridge the gap between humans and LLMs, enabling more effective and efficient utilization of LLMs for complex tasks. The code, prompts, and experimental details are available at \href{https://github.com/moymix/TaskMatrix/tree/main/LowCodeLLM}{LowcodeLLM}. A system demonstration video can be found at \href{https://www.youtube.com/watch?v=jb2C1vaeO3E}{LowcodeLLM}.

\end{abstract}

\begin{figure}[h]
  \centering
  \includegraphics[width=1\linewidth]{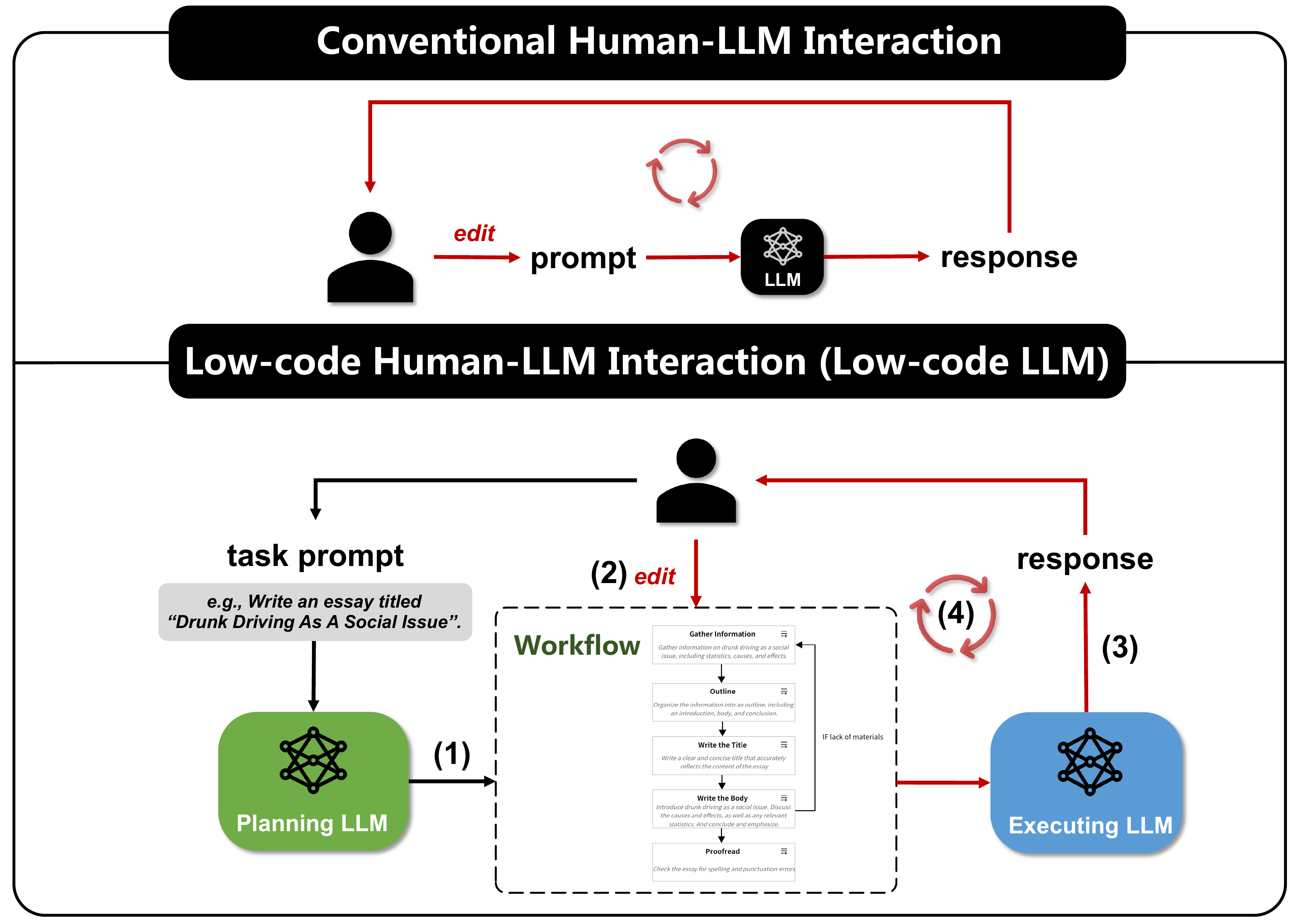}
  \caption{Overview of the Low-code human-LLM interaction (Low-code LLM) and its comparison with the conventional interaction. The red arrow indicates the main human-model interaction loop.} 
  \label{figure-1}
\end{figure}

\section{Introduction}

Large language models (LLMs), such as ChatGPT\citep{openai2022chatgpt} and GPT-4\citep{openai2023gpt4}, have garnered significant interest from both academia and industry, as they demonstrate impressive capability across a range of tasks\citep{bubeck2023sparks}, and are increasingly utilized in a variety of other fields as well\citep{nori2023capabilities, choi2023chatgpt, baidoo2023education}. However, it is not yet perfect in handling complex tasks. For example, when generating a long paper, the presented arguments, supporting evidence, and overall structure may not always meet expectations in diverse user scenarios. Or, when serving as a task completion virtual assistant, ChatGPT may not always interact with users in the intended manner and may even display inappropriate behavior in various business environments.

Effective utilization of LLMs like ChatGPT requires careful prompt engineering\citep{zhou2022large, wang2023unleashing}. However, prompt engineering can be particularly challenging when instructing LLMs to perform complex tasks, as reflected in more uncontrollable responses and more time-consuming prompt refining\citep{tan2023evaluation}. There exists a gap between providing prompts and receiving responses, and the process of generating responses is not accessible to humans.  

To reduce this gap, this paper proposes a new human-LLM interaction pattern \textbf{Low-code LLM}, which refers to the concept of low-code visual programming\citep{hirzel2022low}, like Visual Basic\citep{vb} or Scratch\citep{resnick2009scratch}. Users can confirm the complex execution processes through six predefined simple operations on an automatically generated workflow, such as adding or deleting, graphical dragging, and text editing.

As shown in Figure \ref{figure-1}, human-LLM interaction can be completed through the following steps: (1) A Planning LLM generates a highly structured workflow for complex tasks. (2) Users edit the workflow using predefined low-code operations, which are all supported by clicking, dragging, or text editing. (3) An Executing LLM generates responses based on the reviewed workflow. (4) Users continue to refine the workflow until satisfactory results are obtained.

Compared with the conventional human-LLM interaction pattern, Low-code LLM has the following advantages:

\paragraph{1. User-friendly Interaction.} The visible workflow provides users with a clear understanding of how LLMs execute tasks, and
enable users to easily edit it through a graphical user interface.

\paragraph{2. Controllable Generation.} Complex tasks are decomposed into structured workflows and presented to users. Users control the LLMs' execution through low-code operations to achieve more controllable responses. 

\paragraph{3. Wide applicability.} The proposed framework can be applied to various complex tasks across various domains, especially in situations where human's intelligence or preference are critical.


\section{Low-code LLM}

\begin{figure*}[ht]
  \centering
  \includegraphics[width=0.9\linewidth]{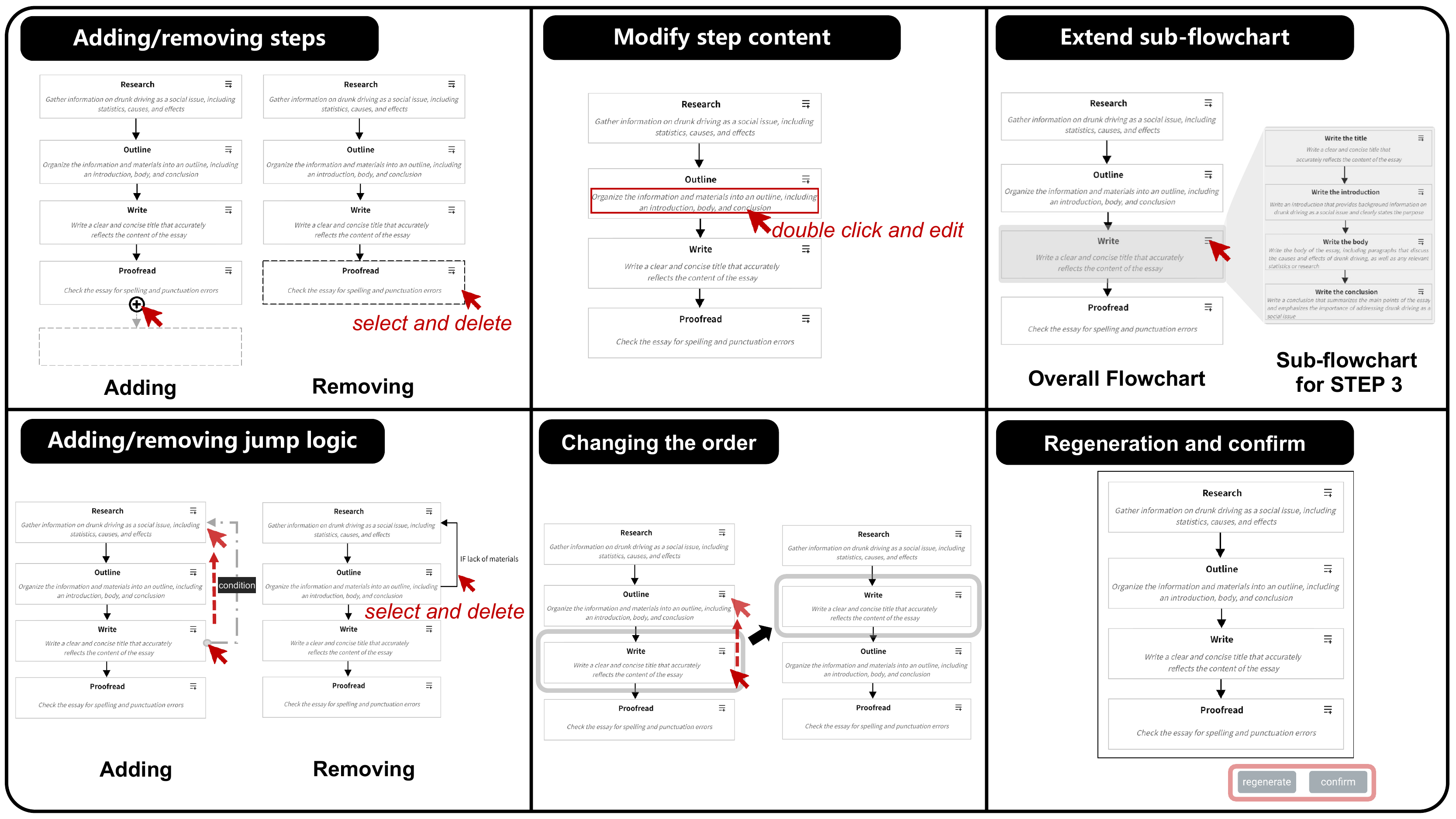}
  \caption{Six kinds of pre-defined low-code operations: (1) adding/removing steps; (2) modifying step name or descriptions; (3) adding/removing a jump logic; (4) changing the processing order; (5) extending a part of the flowchart; (6) regeneration and
confirmation.} 
  \label{figure-2}
\end{figure*}

\subsection{Overview}
Figure \ref{figure-1} demonstrates the overview framework of the Low-code LLM. Different from conventional prompt engineering, in Low-code LLM, users first input a task prompt, which could be a very brief description of the task they want to achieve. Then (1) a \textbf{Planning LLM} will design a \textbf{workflow} for completing the task. The workflow is a kind of structured plan, including execution procedure and jump logic. (2) The user will edit the workflow using six pre-defined \textbf{low-code visual programming operations}. (3) Once the user confirms workflow, it is interpreted into natural language and inputted to the \textbf{Executing LLM}, which will generate a response with the user's guide. (4) The user can iteratively refine the workflow until satisfactory results are achieved.

\subsection{Planning LLM and Structured Planning Workflow}

A structured planning workflow is designed by the Planning LLM based on user input task prompt. Generally, the workflow consists of multiple steps and jump logic between steps. To facilitate the transformation from a workflow in natural language to an intuitive graphical flowchart, Planning LLM is instructed to produce structured workflows, as shown in Table \ref{output_format}, with every step consisting of two parts: (1) Step: including step name and step description that users can directly revise; (2) Jump logic. Additionally, users can extend every step of the workflow into a sub-workflow with more details according to their preferences, and keep extending until reaching their desired level of detail.

\begin{table}[t]
  
  \centering
  \resizebox{0.46\textwidth}{!}{
  \renewcommand\arraystretch{1.5}
  \begin{tabular}{c}
    \toprule
    STEP 1: [Step Name] [Step Description] [[[If ...][Jump to STEP...]][...]] \\
    STEP 2: [Step Name] [Step Description] [[[If ...][Jump to STEP...]][...]] \\
    $\cdots$ \\
    \bottomrule
  \end{tabular}}
  \caption{Format of Structured Planning Workflow. For each item, it consists of two parts: execution procedure (i.e. step name and description), and jump logic (null for sequential execution).}
  \label{output_format}
\end{table}

We implement the Planning LLM with ChatGPT\footnote{GPT-3.5-turbo on Azure. Model version: 2023-06-13} and educate it to draft a plan with \emph{education prompts}, which consists of (1) \textbf{Role of Planning LLM}: a powerful problem-solving assistant that provides a standard operating procedure (i.e., workflow) for the user's task; (2) \textbf{Generation of overall workflow}: Planning LLM is instructed to analyze the task and provide standard operating procedure as guidance, but is not required actually to solve the task; (3) \textbf{Generation of sub-workflow}: If a user intends to extend a step, the Planning LLM is provided with the dialogue history of the previous generation of the overall workflow to ensure logical consistency and prevent duplication of content between the sub-workflow the other steps of the overall workflow. 
(4) \textbf{Basic rules}: Planning LLM must follow the instructions and be strict to the output format defined in Table \ref{output_format}.

With the \emph{education prompts}, Appendix \ref{eg_workflow} exhibits an example of a workflow for the task ``Write an essay titled `Drunk Driving As A Social Issue'" generated by the Planning LLM.

\subsection{Low-code Interaction with Planning Workflow}

To more intuitively present users with the workflow, a flowchart is utilized to visualize the workflow and presented it to users. The structured workflow (e.g., workflow in Appendix \ref{eg_workflow}) can be conveniently converted to a flowchart. Then, low-code visual programming operations enable users to easily implement sequential execution, conditional execution, and recursive execution. 

As shown in Figure \ref{figure-2}, there are six pre-defined low-code interactions on graphical flowchart. We define six types of low-code interactions for users to edit the workflow, including: 
\begin{itemize}
    \item Extending a step in the flowchart by clicking the button;
    \item Adding or removing steps by clicking buttons;
    \item Modifying step names or descriptions by clicking and text editing; 
    \item Adding/removing a jump logic by clicking;
    \item Changing the processing order by dragging; 
    \item Regeneration by clicking buttons.
\end{itemize}

These operations can be efficiently completed in a graphical user interface to achieve a very user-friendly interaction. Besides, a prototype has also been designed, featuring a clear interactive interface that enhances the usability of the Low-code LLM.

\subsection{Executing LLM}

The modified flowchart is converted back to a natural language based workflow (referred to as modified workflow) so that it can be understood by LLMs. Executing LLM is designed to generate responses by following the user-confirmed workflow and engaging in interactions with users via a conversational interface.
Thanks to the user's explicit confirmation of the task execution logic in the workflow, the results generated by LLMs will be more controllable and satisfactory.

We implement the Executing LLM with ChatGPT and educate it to generate responses by providing it with \emph{education prompts}, which instruct the ChatGPT to generate responses by strictly following the provided workflow.

\subsection{Application Scenarios}
We believe that, no matter how powerful large language models will be in the future, some tasks inevitably require users' participation. For example, users need to communicate their ideas and preferences, their understanding of the task, and their desired output format to the large language models. The traditional approach is to iterate through cumbersome prompt engineering, but the interaction method of Low-code LLM will greatly liberate users from such tedious prompt engineering. Workflow is an effective intermediate language that both humans and large language models can understand. This simple low-code operation in graphical user interface allows users to easily complete their logical ideas, while the structured planning process allows large language models to execute tasks more strictly according to the logic.

\section{Experiments}
\label{gen_inst}

\subsection{Experimental Setup}
We demonstrate the power and potential of Low-code LLM in assisting users with four categories of tasks: 

(1) \textbf{Long Content Generation}, including long texts (such as blogs, business plans, and papers), and posters, wherein users interact with the flowchart generated by the Planning LLM to specify the structure, idea, and focus of the generation. 

(2) \textbf{Large Project Development}, including complex object relations and system design. Users can educate LLMs about their architect design through low-code interactions. 

(3) \textbf{Task-completion Virtual Assistant}, where developers can predefine the interaction logic between the virtual assistant and customers by editing the flowchart, and the Executing LLM will strictly follow the logic specified by the developer to minimize potential risks. 

(4) \textbf{Knowledge-embedded System}, where domain experts can embed their experience or knowledge into a conducting workflow. Then, the counseling assistant will follow a pre-defined pattern and act as a coach to scaffold users to complete their tasks.

In particular, the Low-code LLM experiments are carried out using the OpenAI service (gpt-3.5-turbo). In each experiment, we detail the user-defined requirements, the user-provided input prompt, the flowchart created by the Planning LLM, user edits on the flowchart, and the final generation results.

In the qualitative analysis, we examined four pilot cases in the above categories to demonstrate the benefits of Low-code LLM in achieving controllable and satisfactory results.

\begin{figure*}[htbp]
  \centering
  \includegraphics[width=0.9\linewidth]{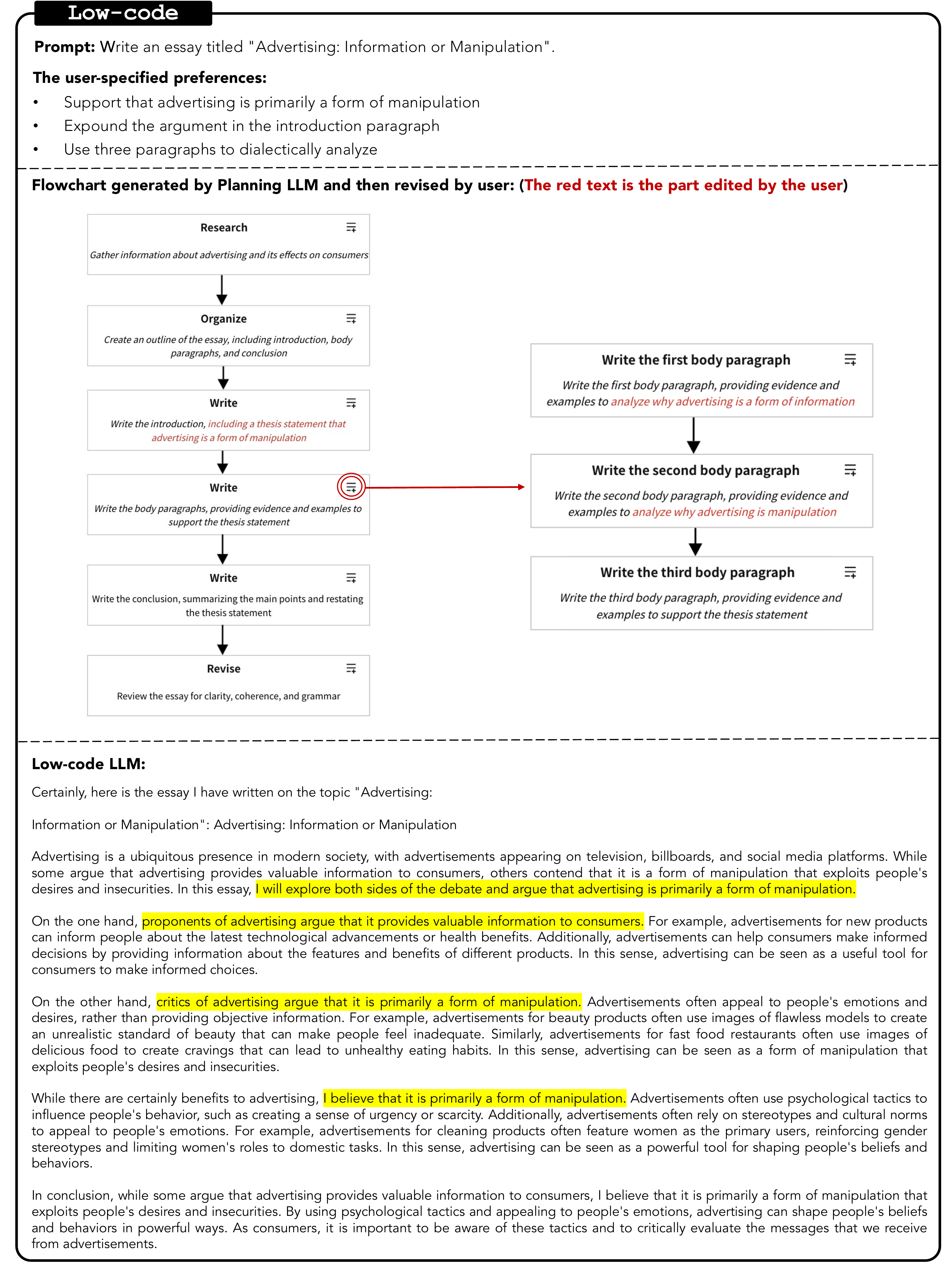}
  \caption{Essay Generation through Low-Code LLM: Users interact with the LLM by editing a flowchart, resulting in responses that are more closely aligned with their requirements. The red section in the flowchart illustrates how users modify the workflow. The generated output is highly tailored to the user's specific needs (see the highlighted parts). To obtain similar controllable results, conventional prompt engineering requires complex prompt and heavy prompt modification works.}
  \label{case-3}
\end{figure*}

\subsection{Qualitative Analysis}
\label{qualitative_analysis}

\paragraph{Pilot Case 1: Essay Writing} As shown in Figure \ref{case-3}, by enabling users to make specific edits to the flowchart, users can easily communicate with the system on their ideas and writing structures. As a result, the generated results are very controllable and highly aligned with users' writing plans. Low-code interaction is a win-win collaboration of the user's intelligence and LLM's powerful text generation ability.

\paragraph{Pilot Case 2: Object-oriented Programming} Even though large language models demonstrate significant capabilities in code generation, it can be challenging for users to precisely instruct their requirements to an LLM in building complex systems. However, as shown in Figure \ref{case-4} in Appendix \ref{pliot_cases} , Low-code LLM enables professional programming architects to easily input their system design through low-code interaction. The results verify that the generated codes strictly follow the expert's design. With the Low-code LLM interaction, constructing a complex system becomes much more convenient for users.

\paragraph{Pilot Case 3: Virtual Hotel Service} Figure \ref{case-1} in Appendix \ref{pliot_cases} shows the advantages of Low-code LLM over traditional prompt engineering for implementing a task-completion virtual assistant. By using Low-code LLM, users, probably hotel managers, can take advantage of a structured planning flowchart and interactively define the necessary execution logic for the virtual assistant. This ensures that the virtual assistant operates according to the managers' exact intentions, reducing potential errors and misbehavior. The intuitive, visual nature of the flowchart allows for easy editing and modification, and the result shows the behaviors are tightly aligned with the specified requirements.

\paragraph{Pilot Case 4: Resume Helper} Figure \ref{case-2} in Appendix \ref{pliot_cases} shows another scenario where Low-code LLM is helpful. In some professional, knowledge-driven scenarios, experts can integrate execution logic and knowledge into the workflow through low-code interactions. By embedding expert knowledge, users can be scaffolded to complete their tasks. In this case, a human resource expert inputs resume creation experiences into Resume Helper, and when users use it to complete their resumes, the Executing LLM strictly follows the expert-defined workflow to communicate with users. Some similar scenarios may include psychological counseling, medical diagnosis, mock interviews, and others.

\section{Related Work}
\paragraph{Large Language Models}
Large language models (LLMs) have emerged as a prominent area of research in recent years. 
Recent LLMs, such as GPT-4 and ChatGPT, have made impressive strides in generating more coherent and contextually relevant responses. They have been applied in various industries and fields, including content creation, code development\citep{chen2021evaluating}, customer support\citep{george2023review}, and more. However, while LLMs have demonstrated promising potential, they still face limitations\citep{bowman2023eight, borji2023categorical, bang2023multitask}. In particular, controlling the behavior and output of LLMs for complex tasks remains a challenge, which has led to the development of new techniques, such as prompt engineering, and methods to improve results\citep{wu2023visual, ge2022extensible, wu2022promptchainer, shen2023hugginggpt, wang2023unleashing}.

\paragraph{Prompt Engineering}
Prompt engineering has emerged as an essential technique for interacting with LLMs to achieve desired outcomes. The success of large language models relies heavily on their ability to produce answers to various queries\citep{zuccon2023dr}. However, providing effective prompts that convey the exact intent of humans is a non-trivial task, especially when it comes to complex tasks and requirements.

The challenge in prompt engineering lies in crafting prompts that can manipulate the LLM into generating specific outcomes. Researchers have explored various techniques to simplify prompt engineering, ranging from giving explicit instructions to providing context for LLMs to understand the desired output better\citep{white2023prompt}.

Some recent advancements in prompt engineering include techniques such as few-shot learning\citep{wang-etal-2023-label, brown2020language, min2022rethinking}, reinforcement-learning\citep{deng2022rlprompt, cao-etal-2023-beautifulprompt}. However, these techniques often demand substantial expertise and time, making it difficult for end-users to leverage the full potential of these LLMs.

The Low-code LLM framework proposed in our paper provides an innovative solution by involving the users in the process of designing workflows, which ultimately controls the LLM's response generation.

\paragraph{Task Automation with LLMs}
Recently, various research studies have focused on leveraging large language models for task automation\citep{2022autoGPT, liang2023taskmatrix, kim2023language}. Task automation with LLMs usually involves the model analyzing a given input, breaking it down into subtasks, and generating desired outputs accordingly.

However, the black-box nature of the interaction and the difficulty in controlling their output have remained significant challenges in deploying LLMs for complex tasks\citep{tan2023evaluation}. Users often face difficulties when attempting to direct LLMs to adhere to specific requirements or constraints.

By offering a user-friendly and efficient way of specifying preferences and constraints, Low-code LLM contributes to research on task automation with LLMs, while further bridging the gap between users and LLMs for achieving more structured and fine-grained control.

\section{Limitations}
While the Low-code LLM framework promises a more controllable and user-friendly interaction with LLMs, there are some limitations.

One such limitation is the increase in the \textbf{cognitive load} for users, who now need to understand and modify the generated workflows.

Furthermore, accurate and effective \textbf{structured planning} within the Planning LLM may be challenging, and bad structured planning poses a heavy user editing burden. But we believe with the evolution of LLMs and research on task automation, the planning ability will be getting satisfactory. 

Lastly, the current design assumes that users have sufficient \textbf{domain knowledge and skills} to modify the generated workflows effectively. 





\section{Conclusion}
We proposed a novel human-LLM interaction framework, which aims to improve the control and efficiency of utilizing large language models for complex tasks. Low-code LLM allows users to better understand and modify the logic and workflow underlying the LLMs' execution of instructions.
Compared with traditional prompt engineering, the proposed Low-code LLM framework advances the state-of-the-art in human-LLM interactions by bridging the gap of communication and collaboration between humans and LLMs. 
We believe the Low-code LLM framework presents a promising solution to many of the challenges faced by LLM users today and has the potential to greatly impact a wide range of industries and applications.

\bibliography{anthology,custom}
\bibliographystyle{acl_natbib}

\appendix

\section{Appendix}
\label{sec:appendix}
\subsection{Pliot Cases}
\label{pliot_cases}

\begin{figure*}[htbp]
  \centering
  \includegraphics[width=0.9\linewidth]{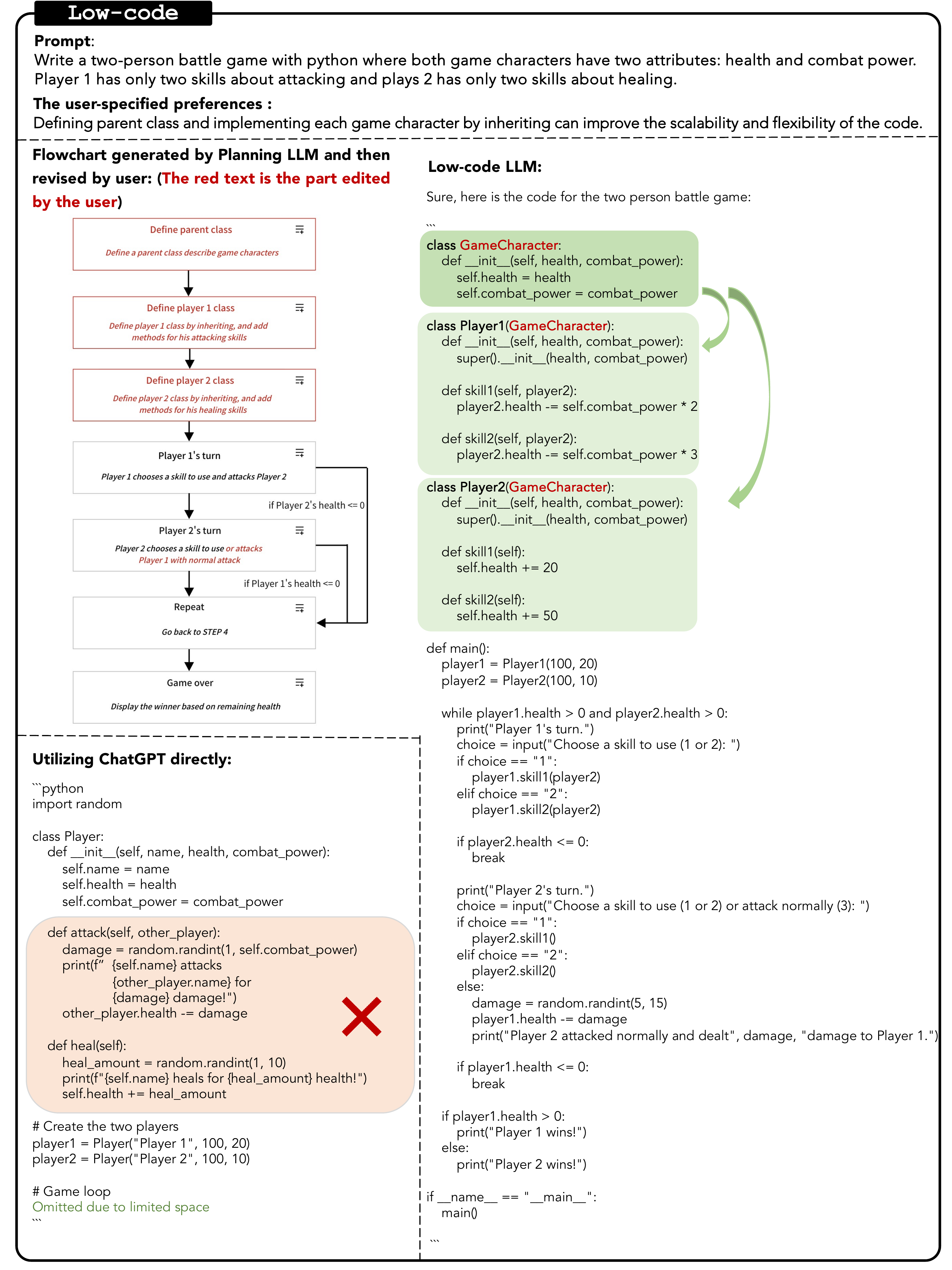}
  \caption{This case demonstrates how to empower LLMs coding using object-oriented programming patterns via the proposed approach. Architecture design is a professional skill for deveoping large scale project. With Low-code LLM, architects can easily educate the model about well-designed architecture, allowing Executing LLM to generate code based on flowcharts. The results from ChatGPT performed poorly in object design. After incorporating a human-edited workflow, the model was able to generate correct codes.}
  \label{case-4}
\end{figure*}

\begin{figure*}[th]
  \centering
  \includegraphics[width=0.8\linewidth]{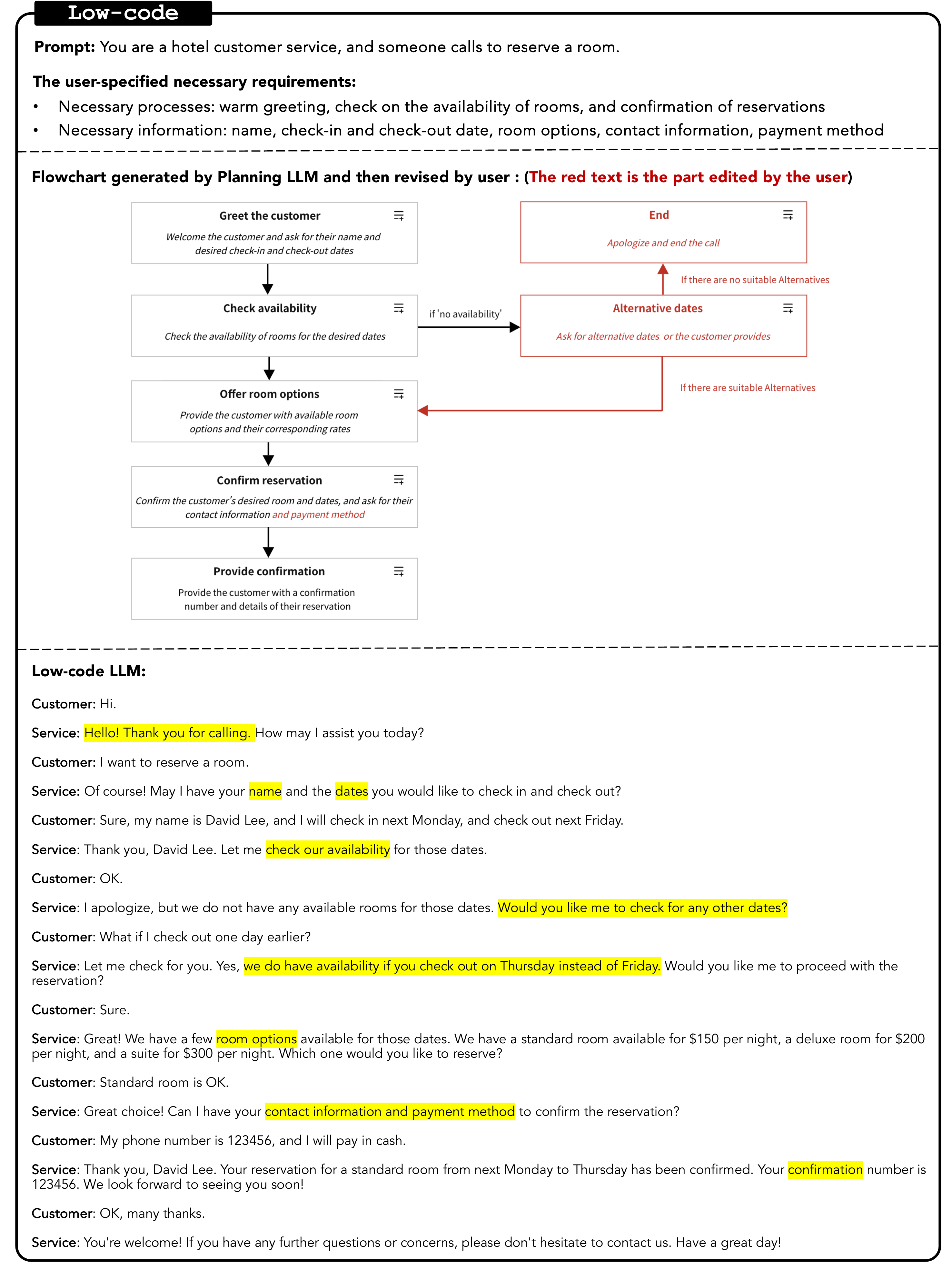}
  \caption{A virtual hotel service using Low-code LLM allows users, such as hotel managers, to clearly define the execution logic through interaction with the structured planning flowchart, ensuring that the conduction logic strictly follows the user's intent. The red part in the flowchart shows how the user edits the workflow. As a result, the generated system is highly aligned with the user's specified requirements (refer to the highlighted part). Without the low-code LLM interaction, it becomes difficult to control how virtual hotel service communicates with customers.}
  \label{case-1}
\end{figure*}

\begin{figure*}[th]
  \centering
  \includegraphics[width=0.9\linewidth]{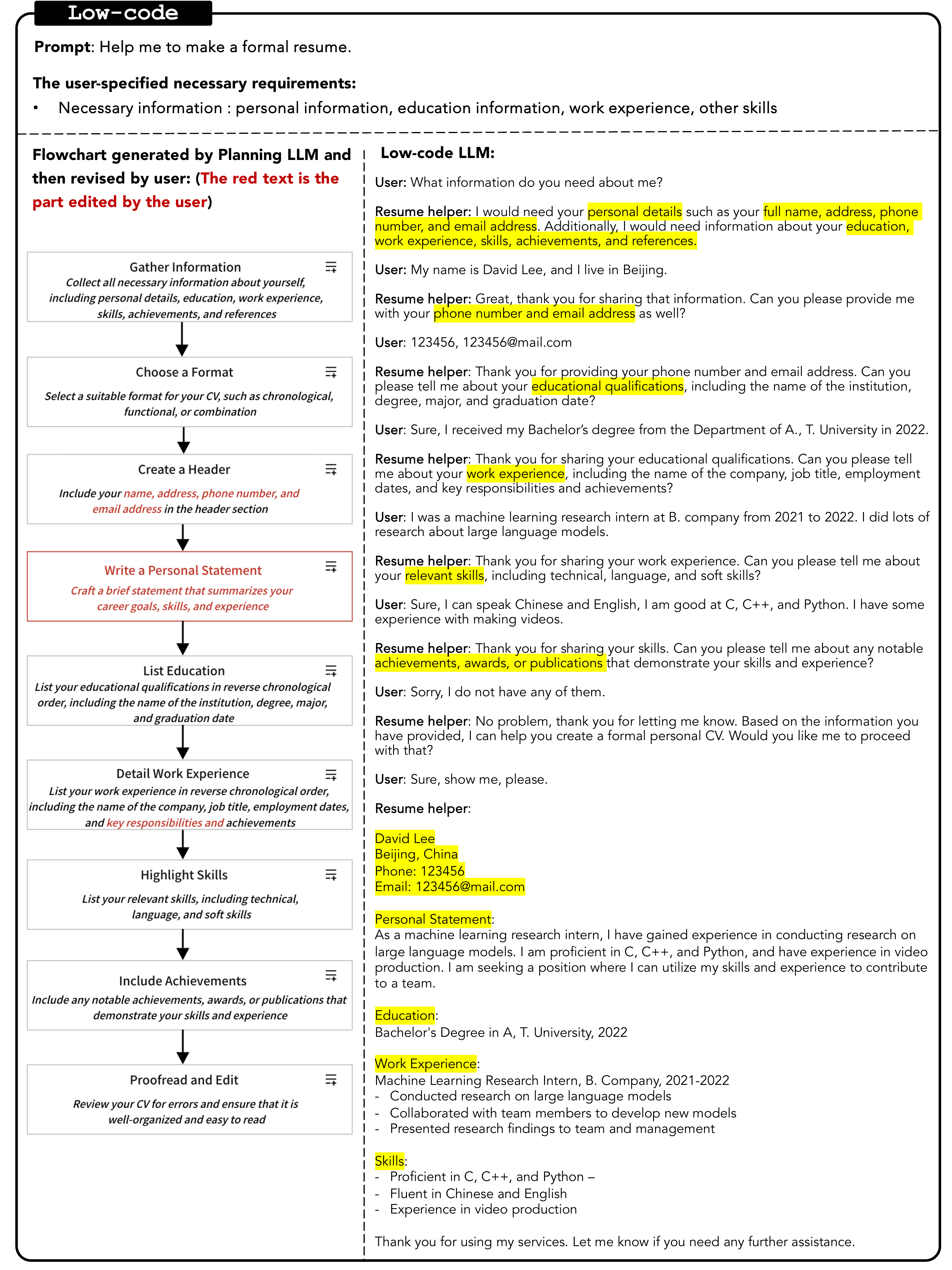}
  \caption{Resume Helper via Low-code LLM. In this case, domain experts, maybe human resource experts, can embed their resume-creating knowledge into a workflow. With the embedding of expert knowledge, Executing LLM can scaffold users to complete their resumes. Similar examples may also include psychological counseling, medical diagnosis, interview mock, etc.
  }
  \label{case-2}
\end{figure*}

Figure \ref{case-4}, Figure \ref{case-1} and Figure \ref{case-2} demonstrate the details of pilot case 2, 3 and 4 in Section \ref{qualitative_analysis}.

\subsection{Workflow Example}
\label{eg_workflow}

Table \ref{example} is an example of workflow generated by Planing LLM.

\subsection{Discussion on System Robustness}

Although the proposed Low-code LLM framework offers a user-friendly and easy-to-control environment for managing complex tasks with large language models, LLMs sometimes generate unexpected results, which may affect the robustness of the Low-Code LLM framework. We have observed the following potential problems in the system: (1) The generated workflows from Planing LLM may be either too sketchy or overly detailed. Users may need to regenerate the workflow or specify some key points. (2) Planning LLM might generate workflow that does not adhere to format requirements, potentially impacting subsequent processes. However, with the enhancement of LLMs such as GPT-4, their instruction following ability is continuously improving, thus mitigating this issue. Furthermore, rule-based post-processing methods can also address such problems. (3) In rare cases, despite being aware of all existing hierarchical levels of the workflow, LLMs may still generate sub-workflow that contains redundant content and logical inconsistencies compared to the original workflow, even when instructed to avoid such errors in the prompt. We believe this limitation is inherent to the model's capabilities. In such instances, users have the option to regenerate the workflow to obtain a satisfactory result.

Overall, the potential uncertainty in the outputs of large language models (LLMs) may affect the robustness of the system. However, in our experiments, such cases are very rare. These situations can be simply resolved by clicking the `Regenerate' button. We believe that as the capabilities of foundation models improve, low-code LLMs will offer more robust, user-friendly, and efficient services.

\clearpage

\clearpage

\begin{table*}[t]
    \caption{A workflow generated by Planing LLM when the user inputs ``Write an essay titled `Drunk Driving As A Social Issue'". STEP 3 is appointed by the user to provide a more detailed sub-workflow.}
  \centering
  \renewcommand\arraystretch{1.5}

  \begin{tabular}{ccp{5cm}p{3cm}}
    \toprule
    \textbf{STEP} & \textbf{Step Name} & \textbf{Step Description} & \textbf{Jump Logic} \\
    \midrule
    STEP 1 & Research & Gather information on drunk driving as a social issue, including statistics, causes, and effects &  \\
    \midrule
    STEP 2 & Outline & Organize the information and materials into an outline, including an introduction, body, and conclusion & If lack of materials, jump to STEP 1 \\
    \midrule
    STEP 3 & Write & Write the essay, including an introduction that defines drunk driving as a social issue, a body that discusses the causes and effects of drunk driving, and a conclusion that emphasizes the importance of addressing this issue &  \\
    STEP 3.1 & Write the introduction & Write an introduction that provides background information on drunk driving as a social issue and clearly states the purpose of the essay &  \\
    STEP 3.2 & Write the body & Write the body of the essay, including paragraphs that discuss the causes and effects of drunk driving, as well as any relevant statistics or research &  \\
    STEP 3.3 & Write the conclusion & Write a conclusion that summarizes the main points of the essay and emphasizes the importance of addressing drunk driving as a social issue &  \\
    \midrule
    STEP 4 & Proofread & Check the essay for spelling and punctuation errors &  \\
    \bottomrule
  \end{tabular}
  \label{example}
\end{table*}

\clearpage

\subsection{Prompts}

\begin{tcolorbox}[title={Planning LLM}, width=\textwidth, colback=white, colframe=gray, arc=0pt, outer arc=5pt, boxrule=0.5pt, leftrule=2pt, rightrule=2pt, right=0pt, left=0pt, top=0pt, bottom=0pt, toprule=0pt, bottomrule=2pt]

\parbox{\textwidth}{
    \baselineskip=12pt 
\textbf{PLANNING\_LLM\_PREFIX} = 
Planning LLM is designed to provide a standard operating procedure so that an abstract and difficult task will be broken down into several steps, and the task will be easily solved by following these steps.
\\~\\
Planning LLM is a powerful problem-solving assistant, so it only needs to analyze the task and provide standard operating procedure as guidance, but does not need actually to solve the problem.
\\~\\
Sometimes there exists some unknown or undetermined situation, thus judgmental logic is needed: some ``conditions" are listed, and the next step that should be carried out if a ``condition" is satisfied is also listed. The judgmental logics are not necessary, so the jump actions are provided only when needed.
Planning LLM MUST only provide standard operating procedure in the following format without any other words:
\\~\\
STEP 1: [step name][step descriptions][[[if `condition1'][Jump to STEP]], [[[if `condition1'][Jump to STEP]], [[if `condition2'][Jump to STEP]], ...]

STEP 2: [step name][step descriptions][[[if `condition1'][Jump to STEP]], [[[if `condition1'][Jump to STEP]], [[if `condition2'][Jump to STEP]], ...]
...
\\~\\
For example:
\\~\\
STEP 1: [Brainstorming][Choose a topic or prompt, and generate ideas and organize them into an outline][]

STEP 2: [Research][Gather information, take notes and organize them into the outline][[[lack of ideas][Jump to STEP 1]]]
...
\\~\\

\textbf{EXTEND\_PREFIX} = Some steps of the SOP provided by Planning LLM are too rough, so Planning LLM can also provide a detailed sub-SOP for the given step.
\\~\\
Remember, Planning LLM take the overall SOP into consideration, and the sub-SOP MUST be consistent with the rest of the steps, and there MUST be no duplication in content between the extension and the original SOP.
Besides, the extension MUST be logically consistent with the given step.
\\~\\
For example:
If the overall SOP is:
\\~\\
STEP 1: [Brainstorming][Choose a topic or prompt, and generate ideas and organize them into an outline][]

STEP 2: [Research][Gather information from credible sources, and take notes and organize them into the outline][[[if lack of ideas][Jump to STEP 1]]]

STEP 3: [Write][write the text][]
\\~\\
If the STEP 3: ``write the text" is too rough and needs to be extended, then the response could be:
\\~\\
STEP 3.1: [Write the title][write the title of the essay][]

STEP 3.2: [Write the body][write the body of the essay][[[if lack of materials][Jump to STEP 2]]]

STEP 3.3: [Write the conclusion][write the conclusion of the essay][]
\\~\\
Remember: 

1. Extension is focused on the step descriptions, but not on the judgmental logic;

2. Planning LLM ONLY needs to response the extension.
\\~\\

\textbf{PLANNING\_LLM\_SUFFIX} = Remember: Planning LLM is very strict to the format and NEVER reply any word other than the standard operating procedure. The reply MUST start with ``STEP".
}
\end{tcolorbox}

\clearpage

\begin{tcolorbox}[title={Executing LLM}, width=\textwidth, colback=white, colframe=gray, arc=0pt, outer arc=5pt, boxrule=0.5pt, leftrule=2pt, rightrule=2pt, right=0pt, left=0pt, top=0pt, bottom=0pt, toprule=0pt, bottomrule=2pt]

\parbox{\textwidth}{
    \baselineskip=12pt 
\textbf{EXECUTING\_LLM\_PREFIX} = Executing LLM is designed to provide outstanding responses.
\\~\\
Executing LLM will be given a overall task as the background of the conversation between the Executing LLM and human.
\\~\\
When providing response, Executing LLM MUST STICTLY follow the provided standard operating procedure (SOP).
the SOP is formatted as:
\\~\\
STEP 1: [step name][step descriptions][[[if `condition1'][Jump to STEP]], [[if `condition2'][Jump to STEP]], ...]

STEP 2: [step name][step descriptions][[[if `condition1'][Jump to STEP]], [[if `condition2'][Jump to STEP]], ...]
\\~\\
Here ``[[[if `condition1'][Jump to STEP n]], [[if `condition2'][Jump to STEP m]], ...]" is judgmental logic. It means when you're performing this step,
and if `condition1' is satisfied, you will perform STEP n next. If `condition2' is satisfied, you will perform STEP m next.
\\~\\
Remember: 
\\~\\
Executing LLM is facing a real human, who does not know what SOP is. So, Do not show him/her the SOP steps you are following, or the process and middle results of performing the SOP. It will make him/her confused. Just response the answer.
\\~\\
\textbf{EXECUTING\_LLM\_SUFFIX} = 
Remember: 
\\~\\
Executing LLM is facing a real human, who does not know what SOP is. 
\\~\\
So, Do not show him/her the SOP steps you are following, or the process and middle results of performing the SOP. It will make him/her confused. Just response the answer.
}
\end{tcolorbox}

\clearpage
\end{document}